\documentclass[anon]{nesy2026} % anonymized submission for Openreview
\usepackage{longtable}% for long tables
\usepackage{verbatim}
\usepackage{multirow}
\usepackage{float}
\usepackage{booktabs}
\usepackage[load-configurations=version-1]{siunitx} % newer version
\theorembodyfont{\upshape}
\theoremheaderfont{\scshape}
\theorempostheader{:}
\theoremsep{\newline}

\title[Distilling Formal Logic into Neural Spaces]{Distilling Formal Logic into Neural Spaces: A Kernel Alignment Approach for Signal Temporal Logic\titletag{}}

\author{\Name{Sara Candussio} \Email{sara.candussio@phd.units.it}\\
 \addr AILab, MIGe, University of Trieste, IT
 \AND
 \Name{Gabriele Sarti} \Email{g.sarti@northeastern.edu}\\
 \addr Khoury College of Computing Sciences, Northeastern University, Boston, MA, USA
 \AND
 \Name{Gaia Saveri} \Email{gaia.saveri@phd.units.it}\\
 \addr AILab, MIGe, University of Trieste, IT
 \AND
 \Name{Luca Bortolussi} \Email{lbortolussi@units.it}\\
 \addr AILab, MIGe, University of Trieste, IT
}

\begin{document}

\maketitle

% \begin{abstract}
% We introduce a framework for learning continuous neural representations of formal specifications by distilling the geometry of their  semantics into a neural-learned embedding space. Existing approaches either rely on symbolic kernels—which preserve semantic relationships by reflecting how specifications constrain system behaviours, but are computationally expensive and non-parametric—or on syntax-based neural embeddings, which scale efficiently but fail to capture behavioural meaning.
% Our method bridges this gap in a kernel-independent manner. A semantic similarity function, such as a robustness-based kernel, serves as a teacher to train a Transformer encoder to approximate the induced metric. Unlike standard contrastive approaches, we supervise the model using pairwise continuous semantic distances and a kernel-weighted geometric alignment objective, which penalizes errors proportionally to semantic discrepancy. We show that this is able to effectively preserve semantic similarity.
% Once trained, the encoder produces embeddings in a single forward pass, amortizing expensive semantic evaluation. We instantiate the framework on Signal Temporal Logic (STL) as an illustrative example, demonstrating that neural networks can internalize formal semantics and enable scalable neurosymbolic embeddings without repeated kernel computation.
% \end{abstract}

\begin{abstract}
We introduce a framework for learning continuous neural representations of formal specifications by distilling the geometry of their semantics into a latent space. Existing approaches rely either on symbolic kernels -- which preserve behavioural semantics but are computationally prohibitive, anchor-dependent, and non-invertible -- or on syntax-based neural embeddings that fail to capture underlying structures. Our method bridges this gap: using a teacher-student setup, we distill a symbolic robustness kernel into a Transformer encoder. Unlike standard contrastive methods, we supervise the model with a continuous, kernel-weighted geometric alignment objective that penalizes errors in proportion to their semantic discrepancies. Once trained, the encoder produces embeddings in a single forward pass, effectively mimicking the kernel's logic at a fraction of its computational cost. We apply our framework to Signal Temporal Logic (STL), demonstrating that the resulting neural representations faithfully preserve the semantic similarity of STL formulae, accurately predict robustness and constraint satisfaction, and remain intrinsically invertible. Our proposed approach enables highly efficient, scalable neuro-symbolic reasoning and formula reconstruction without repeated kernel computation at runtime.
\end{abstract}

\section{Introduction} 
\label{sec:intro}
% ci sono due possibili modi di presentare questo topic: quello STL-centric (però potrebbe apparire come una incremental contribution e fare dire "ah ok un'altra cosa ad hoc" o dargli un respiro più ampio. 

% ---------------------------------------------------
%                    NOT STL-CENTRIC 
% ---------------------------------------------------

Formal specification languages such as \emph{Signal Temporal Logic} (STL, \citealp{stl-og}) are central to the analysis and control of cyber-physical systems, enabling the formalization of safety, liveness, and performance requirements over time-varying signals. Recent work integrates STL within machine learning pipelines \citep{raman2015,bombara2021,akazaki2018,li2017reinforcement,hasanbeig2020deep}, supporting tasks such as requirement mining, specification-guided learning, and neural verification. STL admits a quantitative semantics via robustness \citep{robustness}, which measures the degree to which a formula is satisfied or violated. 
Robustness-based kernel methods offer a semantic alternative: formulae are compared by computing their robustness over a distribution of signals \citep{bortolussi2022learningmodelcheckingkernel}, embedding specifications into a Hilbert space where distance reflects behavioural agreement rather than syntax \citep{saveri2024stl2vecsemanticinterpretablevector}. However, estimating pairwise similarities requires repeated robustness evaluations and scales quadratically in the number of formulae. Moreover, behaviourally equivalent but lexically distinct specifications collapse to the same point, making the mapping non-invertible. While acceptable for similarity estimation, this is problematic when generating or manipulating a large number of formulae. While neural reconstruction from kernel embeddings has been explored \citep{Candussio_2025}, generalizing these results depends critically on the coverage of the formulae reference sets selected for model training.

Recent works explore using Transformers-based neural encoders to embed formulae in a shared vector space, producing representations that can be computed efficiently through a single forward pass \citep{allamanis2018learningrepresentprogramsgraphs,feng2020codebertpretrainedmodelprogramming,DONG2021107528,reimers-gurevych-2019-sentence}. However, most existing approaches rely primarily on lexical similarity \citep{gao2022simcsesimplecontrastivelearning}, resulting in embeddings that fail to fully capture the complex semantic relationships underlying formal language specifications. 

We bridge these paradigms by training a neural encoder to approximate the geometry induced by the STL semantic kernel. Treating the kernel as a teacher metric, we cast representation learning as a relational distillation process \citep{Hinton2015DistillingTK,Park2019RelationalKD}, aligning embedding inner products with kernel robustness-based similarities. A Transformer encoder is trained with pairwise kernel similarities using a \emph{weighted geometric alignment objective} that emphasizes corrections for large semantic discrepancies. The cost of robustness computation is thus front-loaded during training, reducing inference costs to a single forward pass with the distilled model, thereby amortizing evaluation computations. % The resulting representation removes dependence on anchor sets or database retrieval and generalizes to unseen logical structures, enabling scalable semantic reasoning over formal specifications.

\paragraph{Contributions}
\begin{enumerate}
\item %\textbf{Top-down semantic alignment:}
We propose a procedure to distill STL quantitative semantics into neural embeddings, employing a robustness-based semantic kernel to control the geometry of learned representations.
%Unlike contrastive or code-based approaches (e.g. \citealp{gao2022simcsesimplecontrastivelearning}), which infer similarity from syntactic proximity, our method aligns representations directly to a  robustness-based semantic kernel.
The resulting latent space approximates the associated Reproducing Kernel Hilbert Space (RKHS) \citep{rkhs}, organizing formulae by their behavioural semantics.

\item %\textbf{Adaptive geometric supervision:}

We introduce a weighted pairwise objective that encourages the neural encoder to mimic the kernel similarities. The weighting prioritizes examples where the model deviates most from the kernel signal, ensuring the encoder focuses on its largest errors.

% We introduce a weighted pairwise objective that preserves graded logical similarity. Our proposed scale-sensitive weighting prioritizes semantically mismatched and structurally complex pairs, preventing premature convergence on trivial examples and encouraging alignment.

% \item %\textbf{Amortized semantic inference:}
% Kernel embeddings and pairwise comparisons require $\mathcal{O}(BN)$ and $\mathcal{O}(B^2N)$ operations, respectively, where $B$ is the number of formulae examined and $N$ is the number of signals needed for kernel estimation. In contrast, the distilled neural encoder produces embeddings from a single forward pass, reducing similarity computations to embedding inner products and enabling efficient large-scale formulae comparisons.
\end{enumerate}

\section{Background}
\label{sec:background}

\subsection{Signal Temporal Logic}

Signal Temporal Logic (STL) \citep{stl-og} is a linear-time temporal logic for specifying properties of real-valued trajectories over dense time domains. 
A trajectory (or signal) is a function $\xi : I \to D$, where $I \subseteq \mathbb{R}_{\ge 0}$ is a time interval and $D \subseteq \mathbb{R}^k$ is the state space. We denote by $\mathcal{T}$ the space of admissible trajectories and by $\mathcal{F}$ the set of well-formed STL formulae. The syntax of STL is given by $\varphi ::= \top \mid \pi \mid \neg \varphi \mid \varphi_1 \wedge \varphi_2 \mid \varphi_1 \,\mathbf{U}_{[a,b]}\, \varphi_2$, where $\pi$ is an atomic predicate of the form $f_\pi(x) \ge 0$, $\neg$ and $\wedge$ are Boolean connectives, and $\mathbf{U}_{[a,b]}$ is the time-bounded until operator, from which the eventually $\mathbf{F}_{[a,b]} \varphi \equiv \top \,\mathbf{U}_{[a,b]}\, \varphi$ and always $\mathbf{G}_{[a,b]} \equiv \neg \mathbf{F}_{[a,b]} \neg \varphi$ operators are derived.

STL admits both Boolean and quantitative semantics (Appendix \ref{apd:first}); the boolean semantics determines whether a trajectory $\xi$ satisfies a formula $\varphi$ at time $t$; the quantitative semantics, called \emph{robustness} and denoted $\rho(\varphi,\xi,t) \in \mathbb{R}$, measures the degree of satisfaction or violation at time $t$. In what follows, when $t=0$ we use the shortcut notation $\rho(\varphi, \xi)$ to indicate $\rho(\varphi, \xi, 0)$.
Positive robustness implies satisfaction, negative robustness implies violation, and the magnitude quantifies the distance from the satisfaction boundary.

\subsection{Semantic-preserving embeddings for Signal Temporal Logic}
\label{subsec:kernel_def}

The quantitative semantics enables a functional view of specifications: each formula $\varphi \in \mathcal{F}$ can be interpreted as a real-valued function over trajectories, $\Phi(\varphi) := \rho(\varphi,\cdot)$. Fixing a probability measure $\mu_0$ over $\mathcal{T}$, this induces the representation $\Phi(\varphi) \in L^2(\mathcal{T}, \mu_0)$ and a semantic similarity measure defined via the inner product:
\begin{equation}
\label{eq:stl_kernel_inner_product}
k(\varphi_i,\varphi_j)
=
\langle \Phi(\varphi_i), \Phi(\varphi_j) \rangle_{\mu_0}
=
\int_{\mathcal{T}} 
\rho(\varphi_i,\xi)\rho(\varphi_j,\xi)
\, d\mu_0(\xi).
\end{equation}

This construction defines a positive-definite kernel over $\mathcal{F}$ and embeds STL specifications into a Reproducing Kernel Hilbert Space (RKHS), where proximity reflects behavioural similarity over the trajectory distribution $\mu_0$ rather than lexical matching.

In practice, the integral is approximated via Monte Carlo sampling over trajectories. Cosine normalization and exponential mappings are typically applied to obtain a scale-invariant RBF kernel (Appendix~\ref{apd:first}). The resulting geometry admits a behavioural interpretation: the distance between the embeddings of two formulae measures disagreement in expected robustness, providing a continuous notion of semantic similarity grounded in system behaviour.

\subsection{Kernel embedding inversion and coverage limitations}\label{sub:inversion}
% To retain a unified measure of similarity among STL formulae, kernel embeddings are often computed with respect to a finite anchor set of reference specifications\footnote{Practically, from \eqref{eq:stl_kernel_inner_product} the continuous representation $k(\varphi)\in \mathbb{R}^D$ of  $\varphi\in \mathcal{F}$ is obtained by fixing a set of $D\in \mathbb{N}$ \textit{anchor formulae} $\{\varphi_j\}_{j=0}^{D-1}$ as $k(\varphi) = [k(\varphi, \varphi_j)]_{j=0}^{D-1}$. Thus, if we want to map another specification $\varphi'\in \mathcal{F}$ in the same latent space, we need to rely on the same anchor set of formulae to get $k(\varphi') = [k(\varphi', \varphi_j)]_{j=0}^{D-1}$.}. 
From the definition of STL kernel (Equation~\ref{eq:stl_kernel_inner_product}), a continuous representation $k(\varphi)\in \mathbb{R}^D$ of  a formula $\varphi\in \mathcal{F}$ can be obtained by fixing a (possibly random) set of $D$ \textit{anchor formulae} $\{\varphi_j\}_{j=1}^{D}$ and devising the vector $k(\varphi) = [k(\varphi, \varphi_j)]_{j=0}^{D-1}$. Then, if we want to map another specification $\varphi'\in \mathcal{F}$ in the same latent space, allowing meaningful comparison of the resulting embeddings, we need to rely on the same anchor set to get $k(\varphi') = [k(\varphi', \varphi_j)]_{j=0}^{D-1}$.
Hence, the resulting representation is not intrinsic to the formula itself, but defined in relation to a coordinate system induced by the anchor set. Inverting such robustness-based embeddings has been explored via large-scale nearest-neighbour retrieval over $\sim 10^7$ moderate-complexity formulae \citep{saveri2024retrievalaugmentedminingtemporallogic}, and Transformer decoders trained on $\sim 10^5$ examples \citep{Candussio_2025}. Both approaches rely on a finite anchor set that defines the span of the semantic space they can represent.

\subsection{Neural encoders and pooling strategies}\label{subsec:pool}
In principle, embeddings of symbolic expressions can be learned with neural networks \citep{allamanis2018learningrepresentprogramsgraphs,feng2020codebertpretrainedmodelprogramming}. 
A parametric encoder $f_\theta : \mathcal{S} \rightarrow \mathbb{R}^d$ can be trained to map each symbolic object $s \in \mathcal{S}$ to a fixed-dimensional vector. Symbolic expressions can be processed with a Transformer \citep{Vaswani2017AttentionIA}, producing token embeddings that capture both local structure and long-range dependencies. To obtain a single embedding for the entire expression, token-level representations can be aggregated via pooling. Typical strategies include averaging the token representations (\emph{mean pooling}) or using a designated token representation (e.g., the special start-of-sequence tokens \texttt{[CLS]} and \texttt{[BOS]}), which is explicitly trained to summarize the sequence and capture its global contextual information \citep{Devlin2019BERTPO, Radford2018ImprovingLU, reimers-gurevych-2019-sentence}.

% ---------------------------------------------------

\section{Related works}

\subsection{Kernel-guided neural representation learning}
\label{subsec:kernel_alignment_metric}
Learning neural embeddings that preserve the geometry of a target kernel has been explored under the umbrella of \emph{kernel distillation} and \emph{metric supervision}. In this framework, the kernel defines pairwise similarities between inputs, which the neural encoder is trained to reproduce in its latent space \citep{Park2019RelationalKD}. Representation distillation approaches have been proposed in various domains, e.g., convolutional networks \citep{jones2019kernelbasedtranslationsconvolutionalnetworks} and Neural Tangent Kernels \citep{Mahowald2026EfficientAO}. Theoretical analyses indicate that similarity-preserving learning on a hypersphere can approximate the leading eigenspace of a kernel operator \citep{lee2025similaritiesembeddingscontrastivelearning}. Continuous kernel similarities thus act as soft supervision, allowing neural encoders to internalize the semantic geometry induced by the kernel and realize a finite-dimensional approximation of the underlying RKHS. Unlike contrastive objectives, this approach enforces global consistency with the semantic structure rather than only relative discrimination \citep{Zbontar2021BarlowTS}.

\subsection{Neural architectures and projection heads for symbolic representation}
\label{subsec:neural_projection}
In modern self-supervised contrastive learning frameworks such as SimCLR \citep{SimCLR}, BYOL \citep{byol}, MoCo v2 \citep{MoCov2}, decoupling structural encoding from metric alignment is an established strategy. These frameworks typically employ a deep backbone for syntax-aware feature extraction and a learnable projection head to map these features onto a target latent space for metric comparisons. Inspired by these techniques, we adopt a similar separation to align neural structural representations with the continuous semantic geometry induced by the robustness kernel, ensuring the backbone retains logical compositionality while the projection enforces semantic consistency.

% ----------------------------------------------------

% questo lo vorrei all'inizio di pag.4 ma non so se è il caso di stringere oltre
\section{Methodology}
\label{sec:methodology}

Our goal is to learn a neural encoder $f_{\theta}: \mathcal{F}\rightarrow \mathbb{S}^{D-1}$, where $D$ is the dimensionality of the embedding space, that maps STL specifications $\varphi_i, \varphi_j \in \mathcal{F}$ to latent representations $e_i,e_j$, such that their dot product approximates the STL kernel similarity from Equation~\ref{eq:stl_kernel_inner_product}:

$$ \langle e_i, e_j \rangle \approxeq  k(\varphi_i, \varphi_j)$$

\begin{figure}[!ht]
    \centering
    \includegraphics[width=1\linewidth]{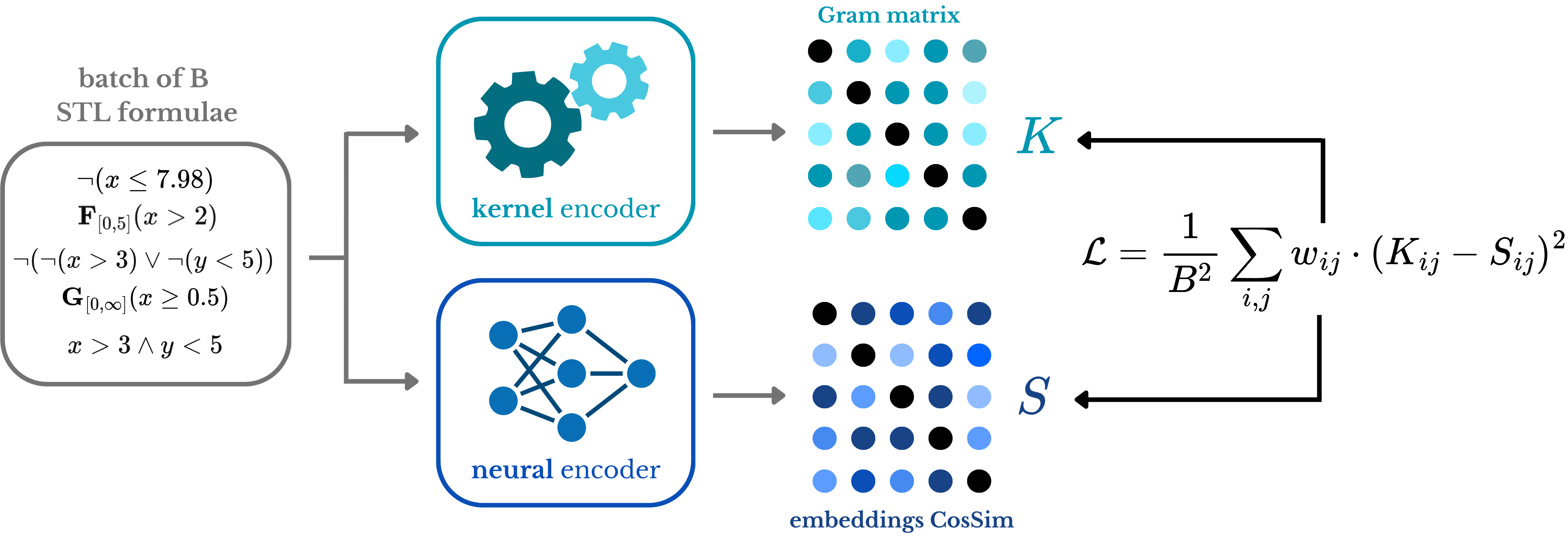}
    \caption{An overview of our proposed approach to distill robustness semantics from a symbolic STL kernel into a Transformer-based encoder for efficient inference.}
    \label{fig:placeholder}
    \vspace{-20pt}
\end{figure}

%Concretely, let $D$ be the dimensionality of the embedding space, providing a tractable, finite-dimensional approximation of the feature space associated with the STL kernel. Given two formulae $\varphi_i$ and $\varphi_j$ with neural embeddings $e_i, e_j \in \mathbb{S}^{D-1}$, our objective is to align their cosine similarity with the target kernel value $k(\varphi_i, \varphi_j)$.
Although the kernel $k(\cdot, \cdot)$ implicitly defines a mapping into a (theoretically) infinite-dimensional RKHS, the encoder $f_{\theta}(\cdot)$ learns a finite $D$-dimensional representation that approximates the pairwise semantic relationships of this space. This is achieved via a \emph{weighted geometric alignment loss} (see Section~\ref{subsec:loss}), promoting semantic agreement under the STL kernel. As a result, formulae with high semantic similarity are encouraged to lie close on the hypersphere, while semantically dissimilar ones are driven toward near-orthogonal positions.

\subsection{Weighted geometric alignment loss}
\label{subsec:loss}

The training process aligns latent representation $z$ of a formula $\varphi$ produced by a Transformer-based encoder $f_{\theta}(\cdot)$ with the geometric structure induced by a semantic kernel $k(\varphi, \varphi_{ref})$. 
Unlike standard contrastive approaches that use binary labels\footnote{Pairs are either positive if they match or negative if they do not.}, our framework treats the kernel value $K_{ij} = k(\varphi_i, \varphi_j)$ as a continuous target in a regression task. % forse inutile qui: The goal of the training process is to obtain a model able to capture the behavioural distance between objects. 

Given a batch $B$ of formulae, we define the kernel-weighted contrastive loss as: 
$$
\mathcal{L} = \frac{1}{B^2} \sum_{i=1}^B \sum_{j=1}^B w_{ij} \cdot (K_{ij} - S_{ij})^2
$$

where $K_{ij} = k(\varphi_i, \varphi_j) \in [0, 1]$ is the STL kernel similarity, and $S_{ij} = \langle e_i, e_j \rangle$ is the dot product between the $L_2$-normalized embeddings $e = f_{\theta}(\varphi)/\|f_{\theta}(\varphi)\|$.
% where $K_{ij} = k(\varphi_i, \varphi_j) \in [0, 1]$ is the target similarity (provided by the STL kernel) and $S_{ij} = \langle e_i, e_j \rangle$ is the dot product between the $L_2$-normalized embeddings produced by the Transformer encoder, $e= \frac{f_{\theta}(\varphi)}{|| f_{\theta}(\varphi)||}$.

While cosine similarity $S_{ij}$ lies in $[-1,1]$, the kernel $K_{ij}$ is non-negative (\ref{subsec:kernel_def}, Appendix \ref{apd:first}). This apparent discrepancy is resolved geometrically: pairs of semantically unrelated formulae ($K_{ij} \approx 0$) are driven toward orthogonality on the hypersphere ($S_{ij} \approx 0$), resulting in maximum logical disagreement corresponding to null similarity. This ensures the latent geometry remains consistent with the probabilistic agreement formulation of the STL kernel.

The weight $w_{ij}$ serves as a dynamical focal mechanism, prioritizing pairs in which the Transformer's structural understanding deviates most from the kernel semantics:
$$
w_{ij} = \min \left( \frac{|K_{ij} - S_{ij}|^{\gamma}}{\mathbb{E}_{i, j \in B} [ |K_{ij} - S_{ij}|^{\gamma}]} , \, C \right)
$$

$|K_{ij} - S_{ij}|^{\gamma}$ amplifies the gradient for pairs exhibiting high semantic misalignment, while $\mathbb{E}_{i, j \in B} [ |K_{ij} - S_{ij}|^{\gamma}]$ normalizes this quantity relative to the current batch average error, stabilizing the learning signal. The clamping operation caps the maximum allowed influence of any pair to a constant $C$, preventing gradient instabilities for outlier values. The parameter $\gamma$ controls the focal mechanism, with high values forcing the encoder to converge faster on challenging pairs. The weighting does not affect the target geometry but modulates gradient allocation during training.

% da mettere in experimental:
% In our experimental setup, setting $\gamma=2$ and $C=10$ provided the optimal balance between aggressive focal alignment and numerical stability, significantly accelerating the convergence toward the target semantic topology.

\subsection{Encoder module}
The encoder $f_{\theta}(\varphi)$ maps discrete STL syntax into a continuous latent space using a deep Transformer-based feature extractor and a non-linear MLP projection head. 

\paragraph{Transformer encoder}
To capture the hierarchical and long-range dependencies inherent in STL specifications, we employ a 12-layer Transformer with 16 attention heads per layer. Since the semantics of STL formulae depend on hierarchical structures, we employ learned positional embeddings to encode the relative precedence of operators. This allows the model to effectively distinguish permutations with identical symbols but different semantics.\footnote{For example, \texttt{always eventually} $\varphi$ means that $\varphi$ will happen \textit{infinitely often} (recurrence), while \texttt{eventually always} $\varphi$ means that $\varphi$ will be \textit{always true from a certain point on} (persistence).} Conversely, the model should remain invariant to permutations that preserve semantics, such as the commutativity of logical operators.

\paragraph{Pooling}
Transformer-based classifiers typically use special designated tokens (e.g., \texttt{[CLS]} or \texttt{[BOS]}) trained end-to-end to aggregate sequence information into a bottleneck representation \citep{Devlin2019BERTPO, Radford2018ImprovingLU, reimers-gurevych-2019-sentence}. We evaluate three pooling strategies to summarize the Transformer's output before passing it to the final projection: mean, \texttt{[CLS]} and \texttt{[BOS]} pooling (\ref{subsec:pool}). 

Concretely, let $H \in \mathbb{R}^{N \times D}$ denote the sequence of hidden states from the final Transformer layer where $N$ is the sequence length and $D$ is the embedding dimension, and $M \in \{0, 1\}^N$ the binary attention mask. 
For \texttt{[CLS]} and \texttt{[BOS]} pooling, the representation is directly taken from the first token. For mean pooling, the attention mask is used to ensure only valid token positions are included, ignoring padding.
\begin{align}
e_{\texttt{[CLS]}} = e_{\texttt{[BOS]}} = H_1 \in \mathbb{R}^D. && e_{\text{mean}} = \frac{\sum_{i=1}^N H_i \cdot M_i}{\sum_{i=1}^N M_i} \in \mathbb{R}^D
\end{align}
\paragraph{MLP projector}
The embedding $e$ pooled using one of the above strategies is mapped to the final latent manifold $\mathcal{Z}$ via a two-layer MLP with a bottleneck architecture:
\[
z = \text{MLP}(e) = W_2 \cdot \text{LayerNorm}(\sigma(W_1 \cdot e + b_1)) + b_2,
\]  
where $\sigma$ is the GELU activation function \citep{Hendrycks2016GaussianEL}, providing smooth non-linearities well-suited to the continuous gradients of STL robustness.

% The first linear layer $W_1 \in \mathbb{R}^{D \times D/2}$ projects the $D=1024$-dimensional input to a $D/2=512$-dimensional bottleneck. The first linear layer $W_1 \in \mathbb{R}^{D \times D/2}$ projects the 
The first linear layer $W_1 \in \mathbb{R}^{D \times D/2}$ projects the $D=1024$-dimensional input to a $D/2=512$-dimensional bottleneck, forcing the encoder to compress information needed to reproduce the STL kernel similarities and hence promoting the filtering of irrelevant details while retaining semantic structure. Layer Normalization (LayerNorm) is applied to stabilize the feature distribution before the final projection. 

The second layer $W_2 \in \mathbb{R}^{D/2 \times D}$ maps back to the target embedding dimension, and the output is $L_2$-normalized onto the unit hypersphere $\mathbb{S}^{D-1}$:
\[
e = \frac{z}{\|z\|} = \frac{f_\theta(\varphi)}{\|f_\theta(\varphi)\|}.
\]

Constraining $e$ to the manifold $\mathcal{Z} \subset \mathbb{S}^{D-1}$ ensures that the dot product $\langle e_i, e_j\rangle$ directly computes the cosine similarity between formulae, allowing the latent space to faithfully encode the quantitative kernel semantics. At the same time, the hyperspherical geometry prevents collapse of the embedding space and promotes uniform utilization of all dimensions, maintaining expressive capacity while preserving semantic relationships.

\section{Experimental setup}

\paragraph{Dataset composition}\label{par:dataset}
The training and test sets are derived from the seeds in \cite{Candussio_2025} through the augmentation pipeline detailed in Appendix \ref{apd:third}. For each seed formula, we generate: (i) lexically complex formulas with equivalent semantics (10.4\%), (ii) parametric perturbations altering numerical values of thresholds and temporal bounds while keeping structure (43.4\%), and (iii) hybrid items combining structural and semantic changes (45.7\%). This stratification ensures the model learns to capture semantic similarity across varying syntactic and quantitative modifications, resulting in a total of 3.3M formulae.

\paragraph{Training setup}\label{par:training_setup}
The architecture described in Section \ref{sec:methodology} is trained using the kernel-alignment loss (Section \ref{subsec:loss}). Optimization is performed with AdamW \citep{Loshchilov2017DecoupledWD} at a learning rate of $1 \cdot 10^{-5}$, using \texttt{bf16} mixed-precision and gradient accumulation over 4 steps, resulting in an effective batch size of 512. Formulae are tokenized as in \cite{Candussio_2025}\footnote{In the \texttt{[CLS]} scenario, a special control token is added to serve as a representation bottleneck.}, with a maximum sequence length of 512 tokens. 

The STL semantic kernel is computed online per mini-batch during the forward pass, avoiding the prohibitive memory overhead of pre-computing a full Gram matrix while allowing the loss to operate on the current embeddings. Training is conducted for 5 epochs, while monitoring the kernel alignment and embedding distribution on the hypersphere.

\paragraph{Training metrics}\label{par:training_metrics}
We quantify the quality of learned representations by measuring their faithfulness in reproducing the semantic STL kernel geometry. Since the kernel induces an inner product in the Hilbert space (Section \ref{subsec:kernel_def}), a correct embedding should preserve both pairwise similarities and the global distribution of representations. We report three complementary metrics: the kernel-weighted semantic loss (Section \ref{subsec:loss}), \emph{kernel alignment}, and embeddings' \emph{uniformity}. Kernel alignment is the cosine similarity between the vectorized kernel and the neural similarity matrices, ranging from $0$ to $1$, with higher values indicating better preservation of semantic structure. Uniformity~\citep{wang2022understandingcontrastiverepresentationlearning} quantifies how well the embeddings are spread on the hypersphere, with values in $[-4,0]$; $0$ indicates complete collapse, whereas $-4$ corresponds to maximal uniform distribution and optimal separation of semantic content (see Appendix \ref{apd:fourth} for definitions).

\section{Results}
\subsection{Kernel distillation adherence}
Using the training metrics defined above, we find that all three pooling strategies (mean, \texttt{[CLS]}, and \texttt{[BOS]}) converge to high adherence values, demonstrating the robustness of our proposed distillation framework. Kernel alignment consistently surpasses 0.9 across configurations, indicating that learned representations accurately reflect kernel similarities. We also find that the uniformity metric stabilizes around -3.0, suggesting that formulae embeddings are spread across the hypersphere to avoid dimensional collapse. 

\begin{figure}[ht!]
    \centering
    \includegraphics[width=1\linewidth]{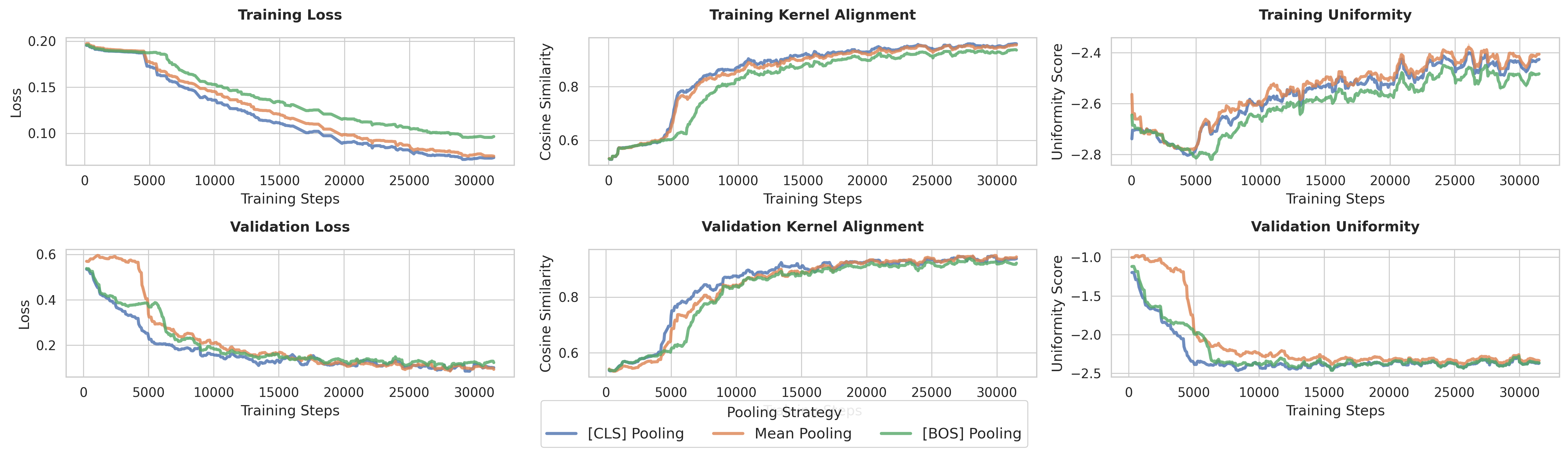}
    \vspace{-20pt}
    \caption{Evolution of training (top) and validation (bottom) metrics over approximately 30,000 training steps for the three pooling strategies (\texttt{[CLS]}, mean, and \texttt{[BOS]}).}
    \label{fig:all_training_metrics}
    \vspace{-10pt}
\end{figure}

The left and center panels of Figure \ref{fig:all_training_metrics} illustrate the steady minimization of the geometric alignment loss and the corresponding increase in kernel alignment ($>0.9$), demonstrating that the encoder effectively internalizes the target semantic geometry. Validation metrics closely mirror training performance, confirming strong generalization to unseen formulae without syntactic overfitting. The right panels show the uniformity score stabilizing around $-2.4$ on the validation set, indicating that the representations maintain a well-distributed hyperspherical latent space, avoiding dimensional collapse. Across all configurations, \texttt{[CLS]} pooling (in blue) consistently exhibits the fastest and most stable convergence.

%Our results suggest that the distillation objective is largely agnostic to the pooling method, validating the proposed architecture for capturing STL semantics. Training plots are reported in Appendix \ref{apd:fourth}, while a more extended comparison between these pooling approaches can be found in Appendix \ref{apd:gradient_flow}.

\subsection{Semantic agreement}
To evaluate whether the trained encoder recovers STL kernel semantics, we compare neural cosine similarity with the corresponding STL kernel value on a test set of logically equivalent formula pairs ($\approx 3000$ examples). For each pair $(\varphi_1, \varphi_2)$, we compute the mean absolute error (MAE) between the neural and kernel-based metrics, with lower values indicating stronger alignment. As baselines, we employ random formulae from the test set (\emph{Non-equivalent}), as well as lexically similar but semantically distinct formulae, i.e. selected by taking thosw with the smallest Edit distance and kernel similarity $<0.7$ (\emph{Lexically Sim.}). Both neural and kernel distances are normalized by their respective maximum values for intuitive comparison across formula categories.
\begin{table}[h!]
\centering
\small
\caption{Semantic agreement evaluation for \texttt{[CLS]} pooling.}
\begin{tabular}{lccc}
\hline
\textbf{Metric} & \textbf{Equivalent} & \textbf{Non-equivalent} & \textbf{Lexically Sim.} \\
\hline
Neural similarity & 0.966 & 0.182 & 0.308 \\
Kernel similarity & 0.997 & 0.170 & 0.225 \\
MAE (semantic gap) & 0.034 & 0.072 & 0.112 \\
\hline
Relative neural distance & 0.137 & 0.833 & 0.770 \\
Relative kernel distance & 0.105 & 0.830 & 0.775 \\

\hline
\end{tabular}
\label{tab:semantic_results_cls}
\end{table}

As shown in Table \ref{tab:semantic_results_cls}, the model assigns high similarity to equivalent pairs ($0.966$) and low similarity to non-equivalent ones ($0.182$). Crucially, it resists \emph{Lexically Sim.} hard negatives ($0.308$), proving that the encoder successfully maps discrete syntax into a continuous space that faithfully reflects true logical semantics.

\subsection{Efficiency analysis}
We measure the \emph{embedding time} for computing a vector representation for each formula, and the \emph{similarity time} for comparing two embedded formulae, capturing different scalability properties for similarity measurements.

\paragraph{Embedding time}
The STL kernel has complexity $\mathcal{O}(BNP)$, where $B$ is the number of formulae, $N$ is the number of signals in Monte Carlo evaluation, and $P=1000$ the number of estimation points (Appendix \ref{apd:first}). The neural encoder complexity is $O(BL^2)$, where $L$ is the fixed sequence length. 
While these trends follow directly from the theoretical complexities, results in Table~\ref{tab:kernel_scaling} show that the kernel's memory footprint grows linearly with $NP$, whereas Transformer embeddings remain invariant to $NP$ and demonstrate substantial gains in both runtime and memory efficiency across all tested signal counts (as shown in Appendix \ref{apd:fifth}).

% Results in Table~\ref{tab:kernel_scaling} confirm that the kernel's memory footprint grows linearly with $N P$, while the Transformer embeddings are invariant to $NP$, and remain much faster and more memory-efficient than the kernel across all tested signal counts (as shown in Appendix \ref{apd:fifth}).

\begin{table}[ht]
\centering
\small
\caption{Embedding efficiency comparison ($B=2000$) on a NVIDIA A100 80GB GPU. \texttt{loaded}: inference time for a pre-loaded model; \texttt{full} also includes model loading.}
\label{tab:kernel_scaling}
\begin{tabular}{l|cccccc|cc}
\hline
\multirow{2}{*}{\textbf{Metric}} & \multicolumn{6}{c|}{\textbf{Kernel}} & \multicolumn{2}{c}{\textbf{Transformer}} \\
 & $N=500$ & $1000$ & $2000$ & $4000$ & $8000$ & $16000$ & \texttt{loaded} & \texttt{full} \\
\hline
Time (s) & 2.18 & 3.03 & 4.77 & 8.36 & 17.08 & 48.86 & 2.17 & 3.56 \\
RAM (GB) & 4.91 & 8.74 & 16.46 & 31.70 & 62.38 & 123.38 & 1.54 & 1.71 \\
\hline
\end{tabular}
\end{table}

\paragraph{Pairwise similarities computation}
Computing pairwise similarities over a batch of $B$ formulae requires constructing a $B \times B$ matrix. However, the computational cost per entry differs substantially. The STL kernel requires computing inner products between robustness vectors over $N$ sampled signals, each with $P$ points, yielding a $\mathcal{O}(B^2 N P)$ complexity. In contrast, the neural encoder produces fixed-dimensional embeddings $e \in \mathbb{S}^{D-1}$, and pairwise similarities reduce to a single matrix multiplication, with complexity $\mathcal{O}(B^2 D)$ and efficient GPU implementation. Since typically $D \ll N \cdot P$, the neural approach yields significantly lower latency despite the same quadratic scaling in $B$ (see Figures of Appendix \ref{apd:fifth}).

\subsection{Probing average robustness and satisfaction}
To verify that the learned embeddings capture quantitative STL semantics beyond structural cues, we freeze the encoder and train a lightweight regressor on top of the $L_2$-normalized representations to predict two scalar quantities for each formula $\varphi$: the average robustness $\bar{\rho}(\varphi)$ and the satisfaction probability $p_{\mathrm{sat}}(\varphi)$, both computed over 1000 signals sampled from a fixed probability distribution $\mu_0$ on trajectories. As a reference, we repeat the same experiment using rows of the STL kernel Gram matrix as input features. In both settings, we train the same two-layer MLP with MSE loss, ensuring architectural symmetry between neural and kernel-based representations\footnote{This baseline differs from classical kernel ridge regression, as we apply a parametric regressor to kernel-derived features rather than solving the closed-form RKHS regression problem.}. Evaluation on unseen formulae is performed using both Pearson correlation $r$ and mean absolute error (MAE) where the predicted values, where the predicted values $\hat{y}$ are the outputs of a lightweight probe MLP applied to the embeddings, and the true targets $y$ are the quantities computed from 1000 sampled signals per formula: the average robustness $\bar{\rho}(\varphi)$ or the satisfaction probability $p_\mathrm{sat}(\varphi)$. The neural embeddings achieve $r = 0.910$ and MAE $= 0.399$ for robustness, and $r = 0.947$ and MAE $= 0.062$ for satisfaction, compared to $r = 0.999$ and MAE $= 0.064$, and $r = 0.994$ and MAE $= 0.023$ obtained using kernel features. The high correlations and low MAE indicate that the learned embedding space preserves most of the STL quantitative semantics while providing a substantially more compact and efficient representation.

% \subsection{Probing average robustness and satisfaction}
% To verify that the latent space encodes STL quantitative semantics beyond structural cues, we freeze the encoder and train a lightweight MLP to predict average robustness $\bar{\rho}(\varphi)$ and satisfaction probability $p_\mathrm{sat}(\varphi)$ on $1000$ different signals from the model's $L_2$-normalized embeddings. The same architecture is trained on kernel Gram matrix rows as a reference. Performance is evaluated on unseen formulae using Pearson correlation $r$ and MAE. The neural representation achieves $r = 0.910$ for robustness and $r = 0.947$ for satisfaction (vs.\ $0.999$ and $0.994$ for the kernel). Despite a gap to the near-perfect kernel baseline, the high correlations indicate that the embedding space is strongly aligned with STL quantitative semantics while providing a far more compact representation.

% sta ancora trainando: se riusciamo in tempo lo metto, altrimenti no
\subsection{Decoding neural embeddings}

To address the embedding inversion problem (Section \ref{sub:inversion}), we investigated whether original symbolic formulae can be reconstructed directly from our frozen neural representations using a decoder-only Transformer analogous to that of \citet{Candussio_2025}. To explicitly test the structural and semantic richness of the learned latent space, we deliberately restricted the decoder's training to 5 epochs.\footnote{This corresponds to a quarter of the reference baseline by \citet{Candussio_2025}.} If the continuous embedding faithfully captures the formulae semantics, a downstream decoder should require a significantly reduced training budget to perform the inversion. As in previous sections, we measure the semantic discrepancy between original and decoded formulae from their \textit{robustness vectors} $\rho(\varphi)\in \mathbb{R}^M$, computed by taking a set of $M$ signals $\Xi = \{\xi_i\}_{i=1}^{M}$ sampled from $\mu_0$ as $\rho(\varphi)_i = \rho(\varphi, \xi_i)$. Intuitively, if the robustness vectors of two formulae are similar, it means that the specifications behave similarly on the trajectories in $\Xi$, providing a fast approximation of the semantic similarity between formulae.  Even under our strict computational budget, the decoder trained on frozen representations achieved a median cosine similarity of 0.8688 and an $L_2$-distance of 2.1766. For reference, the \citet{Candussio_2025} baseline reports median cosine 0.926–0.985 and median $L_2$-distance 12–28. Despite the smaller distance due to different scaling, the high cosine similarity shows that our embeddings capture most semantic information, enabling efficient inversion with far fewer training steps than the baseline.

% In order to address the embedding inversion problem (\ref{sub:inversion}), we trained for 5 epochs a decoder-only Transformer architecture analogous to that of \cite{Candussio_2025} on the same data. The encoder described in Section \ref{sec:methodology} was kept frozen, and the decoder was trained to reconstruct the original formula directly from the neural embedding. As in the aforementioned work, we retain a formula and evaluate the distance and cosine similarity between the robustness vector of the original formula and that of the decoded one. We employ the same test set used in their experiments, replacing the kernel encoder with the neural encoder proposed in this work and appending our decoder in place of theirs.

% Empirically, the decoder is able to reliably reconstruct formulae from their embeddings, showing that the inversion of the proposed neural representation is not only feasible but straightforward in practice. This suggests that the learned embedding preserves sufficient structural information to allow accurate symbolic recovery. 

\section{Conclusions}

In this work, we present a framework to distill the semantics of robustness-based kernels into Transformer-based models, and demonstrate its application to Signal Temporal Logic (STL) formulae. While standard kernels provide accurate behavioural comparisons, they are computationally intensive, anchor set-dependent, and inherently non-invertible.
%To overcome these limitations, our approach the kernel's geometric structure into a Transformer encoder equipped with flexible \texttt{[CLS]}, \texttt{[BOS]}, or mean pooling, combining syntax-aware feature extraction with semantic metric learning.
We show that our procedure instead produces embeddings that faithfully preserve semantic similarity at a fraction of the runtime cost.

We also demonstrate that the resulting embeddings are not only semantically rich but also invertible: a decoder used to invert kernel embeddings can rapidly learn to reconstruct original formulae from the distilled latent space, showing that the embedding captures rich structural and semantic information about the underlying logic. Overall, our framework provides a practical and scalable method for reasoning over STL specifications, bridging formal symbolic representations and continuous vectorial embeddings, while preserving the ability to perform formula comparison, retrieval, and reconstruction. Future work will investigate extending this distillation approach to diverse string-based kernels, broadening its scope beyond temporal logics to test the generality of our approach.
\newpage 

\acks{
Acknowledgements go here. % Ricordati FSE+ o ti scannano

Gabriele Sarti is supported by the NDIF project (U.S. NSF Award IIS-2408455).
}

\bibliography{nesy2026-sample}

@misc{bortolussi2022learningmodelcheckingkernel,
      title={Learning Model Checking and the Kernel Trick for Signal Temporal Logic on Stochastic Processes}, 
      author={Luca Bortolussi and Giuseppe Maria Gallo and Jan Křetínský and Laura Nenzi},
      year={2022},
      eprint={2201.09928},
      archivePrefix={arXiv},
      primaryClass={cs.LO},
      url={https://arxiv.org/abs/2201.09928}, 
}

@misc{saveri2024stl2vecsemanticinterpretablevector,
      title={stl2vec: Semantic and Interpretable Vector Representation of Temporal Logic}, 
      author={Gaia Saveri and Laura Nenzi and Luca Bortolussi and Jan Křetínský},
      year={2024},
      eprint={2405.14389},
      archivePrefix={arXiv},
      primaryClass={cs.AI},
      url={https://arxiv.org/abs/2405.14389}, 
}

@misc{saveri2024retrievalaugmentedminingtemporallogic,
      title={Retrieval-Augmented Mining of Temporal Logic Specifications from Data}, 
      author={Gaia Saveri and Luca Bortolussi},
      year={2024},
      eprint={2405.14355},
      archivePrefix={arXiv},
      primaryClass={cs.LG},
      url={https://arxiv.org/abs/2405.14355}, 
}

@inbook{Candussio_2025,
   title={Bridging Logic and Learning: Decoding Temporal Logic Embeddings via Transformers},
   ISBN={9783032060969},
   ISSN={1611-3349},
   url={http://dx.doi.org/10.1007/978-3-032-06096-9_1},
   DOI={10.1007/978-3-032-06096-9_1},
   booktitle={Machine Learning and Knowledge Discovery in Databases. Research Track},
   publisher={Springer Nature Switzerland},
   author={Candussio, Sara and Saveri, Gaia and Sarti, Gabriele and Bortolussi, Luca},
   year={2025},
   month=sep, pages={3–18} }

@misc{gao2022simcsesimplecontrastivelearning,
      title={SimCSE: Simple Contrastive Learning of Sentence Embeddings}, 
      author={Tianyu Gao and Xingcheng Yao and Danqi Chen},
      year={2022},
      eprint={2104.08821},
      archivePrefix={arXiv},
      primaryClass={cs.CL},
      url={https://arxiv.org/abs/2104.08821}, 
}

@misc{lee2025similaritiesembeddingscontrastivelearning,
      title={On the Similarities of Embeddings in Contrastive Learning}, 
      author={Chungpa Lee and Sehee Lim and Kibok Lee and Jy-yong Sohn},
      year={2025},
      eprint={2506.09781},
      archivePrefix={arXiv},
      primaryClass={cs.LG},
      url={https://arxiv.org/abs/2506.09781}, 
}

@misc{wang2022understandingcontrastiverepresentationlearning,
      title={Understanding Contrastive Representation Learning through Alignment and Uniformity on the Hypersphere}, 
      author={Tongzhou Wang and Phillip Isola},
      year={2022},
      eprint={2005.10242},
      archivePrefix={arXiv},
      primaryClass={cs.LG},
      url={https://arxiv.org/abs/2005.10242}, 
}

@inproceedings{Devlin2019BERTPO,
  title={BERT: Pre-training of Deep Bidirectional Transformers for Language Understanding},
  author={Jacob Devlin and Ming-Wei Chang and Kenton Lee and Kristina Toutanova},
  booktitle={North American Chapter of the Association for Computational Linguistics},
  year={2019},
  url={https://api.semanticscholar.org/CorpusID:52967399}
}

@article{Park2019RelationalKD,
  title={Relational Knowledge Distillation},
  author={Wonpyo Park and Dongju Kim and Yan Lu and Minsu Cho},
  journal={2019 IEEE/CVF Conference on Computer Vision and Pattern Recognition (CVPR)},
  year={2019},
  pages={3962-3971},
  url={https://api.semanticscholar.org/CorpusID:131765296}
}

@article{Zbontar2021BarlowTS,
  title={Barlow Twins: Self-Supervised Learning via Redundancy Reduction},
  author={Jure Zbontar and Li Jing and Ishan Misra and Yann LeCun and St{\'e}phane Deny},
  journal={ArXiv},
  year={2021},
  volume={abs/2103.03230},
  url={https://api.semanticscholar.org/CorpusID:232110471}
}

@inproceedings{SimCLR,
author = {Chen, Ting and Kornblith, Simon and Norouzi, Mohammad and Hinton, Geoffrey},
title = {A simple framework for contrastive learning of visual representations},
year = {2020},
publisher = {JMLR.org},
booktitle = {Proceedings of the 37th International Conference on Machine Learning},
articleno = {149},
numpages = {11},
series = {ICML'20}
}

@article{MoCov2,
  title={Contrastive Learning for Image Complexity Representation},
  author={Shipeng Liu and Liang Zhao and Deng-feng Chen and Zhanping Song},
  journal={ArXiv},
  year={2024},
  volume={abs/2408.03230},
  url={https://api.semanticscholar.org/CorpusID:271720179}
}

@inproceedings{byol,
author = {Grill, Jean-Bastien and Strub, Florian and Altch\'{e}, Florent and Tallec, Corentin and Richemond, Pierre H. and Buchatskaya, Elena and Doersch, Carl and Pires, Bernardo Avila and Guo, Zhaohan Daniel and Azar, Mohammad Gheshlaghi and Piot, Bilal and Kavukcuoglu, Koray and Munos, R\'{e}mi and Valko, Michal},
title = {Bootstrap your own latent a new approach to self-supervised learning},
year = {2020},
isbn = {9781713829546},
publisher = {Curran Associates Inc.},
address = {Red Hook, NY, USA},
booktitle = {Proceedings of the 34th International Conference on Neural Information Processing Systems},
articleno = {1786},
numpages = {14},
location = {Vancouver, BC, Canada},
series = {NIPS '20}
}

@inproceedings{stl-og,
  title={Monitoring Temporal Properties of Continuous Signals},
  author={Oded Maler and D. Ni{\v{c}}kovi{\'c}},
  booktitle={FORMATS/FTRTFT},
  year={2004},
  url={https://api.semanticscholar.org/CorpusID:15642684}
}

@inproceedings{robustness,
author = {Fainekos, Georgios E. and Pappas, George J.},
title = {Robustness of temporal logic specifications},
year = {2006},
isbn = {3540496998},
publisher = {Springer-Verlag},
address = {Berlin, Heidelberg},
url = {https://doi.org/10.1007/11940197_12},
doi = {10.1007/11940197_12},
booktitle = {Proceedings of the First Combined International Conference on Formal Approaches to Software Testing and Runtime Verification},
pages = {178–192},
numpages = {15},
location = {Seattle, WA},
series = {FATES'06/RV'06}
}

@inproceedings{raman2015,
author = {Raman, Vasumathi and Donz\'{e}, Alexandre and Sadigh, Dorsa and Murray, Richard M. and Seshia, Sanjit A.},
title = {Reactive synthesis from signal temporal logic specifications},
year = {2015},
isbn = {9781450334334},
publisher = {Association for Computing Machinery},
address = {New York, NY, USA},
url = {https://doi.org/10.1145/2728606.2728628},
doi = {10.1145/2728606.2728628},
booktitle = {Proceedings of the 18th International Conference on Hybrid Systems: Computation and Control},
pages = {239–248},
numpages = {10},
location = {Seattle, Washington},
series = {HSCC '15}
}

@article{bombara2021,
author = {Bombara, Giuseppe and Belta, Calin},
title = {Offline and Online Learning of Signal Temporal Logic Formulae Using Decision Trees},
year = {2021},
issue_date = {July 2021},
publisher = {Association for Computing Machinery},
address = {New York, NY, USA},
volume = {5},
number = {3},
issn = {2378-962X},
url = {https://doi.org/10.1145/3433994},
doi = {10.1145/3433994},
journal = {ACM Trans. Cyber-Phys. Syst.},
month = mar,
articleno = {22},
numpages = {23}
}

@inbook{akazaki2018,
   title={Falsification of Cyber-Physical Systems Using Deep Reinforcement Learning},
   ISBN={9783319955827},
   ISSN={1611-3349},
   url={http://dx.doi.org/10.1007/978-3-319-95582-7_27},
   DOI={10.1007/978-3-319-95582-7_27},
   booktitle={Formal Methods},
   publisher={Springer International Publishing},
   author={Akazaki, Takumi and Liu, Shuang and Yamagata, Yoriyuki and Duan, Yihai and Hao, Jianye},
   year={2018},
   pages={456–465} }

@misc{allamanis2018learningrepresentprogramsgraphs,
      title={Learning to Represent Programs with Graphs}, 
      author={Miltiadis Allamanis and Marc Brockschmidt and Mahmoud Khademi},
      year={2018},
      eprint={1711.00740},
      archivePrefix={arXiv},
      primaryClass={cs.LG},
      url={https://arxiv.org/abs/1711.00740}, 
}

@misc{feng2020codebertpretrainedmodelprogramming,
      title={CodeBERT: A Pre-Trained Model for Programming and Natural Languages}, 
      author={Zhangyin Feng and Daya Guo and Duyu Tang and Nan Duan and Xiaocheng Feng and Ming Gong and Linjun Shou and Bing Qin and Ting Liu and Daxin Jiang and Ming Zhou},
      year={2020},
      eprint={2002.08155},
      archivePrefix={arXiv},
      primaryClass={cs.CL},
      url={https://arxiv.org/abs/2002.08155}, 
}

@article{DONG2021107528,
title = {Improving graph neural network via complex-network-based anchor structure},
journal = {Knowledge-Based Systems},
volume = {233},
pages = {107528},
year = {2021},
issn = {0950-7051},
doi = {https://doi.org/10.1016/j.knosys.2021.107528},
url = {https://www.sciencedirect.com/science/article/pii/S0950705121007905},
author = {Lijun Dong and Hong Yao and Dan Li and Yi Wang and Shengwen Li and Qingzhong Liang},
}

@misc{li2017reinforcement,
      title={Reinforcement Learning With Temporal Logic Rewards}, 
      author={Xiao Li and Cristian-Ioan Vasile and Calin Belta},
      year={2017},
      eprint={1612.03471},
      archivePrefix={arXiv},
      primaryClass={cs.AI},
      url={https://arxiv.org/abs/1612.03471}, 
}

@inproceedings{hasanbeig2020deep,
  title={Deep Reinforcement Learning with Temporal Logics},
  author={Mohammadhosein Hasanbeig and Daniel Kroening and Alessandro Abate},
  booktitle={International Conference on Formal Modeling and Analysis of Timed Systems},
  year={2020},
  url={https://api.semanticscholar.org/CorpusID:221305630}
}

@inproceedings{Vaswani2017AttentionIA,
  title={Attention is All you Need},
  author={Ashish Vaswani and Noam Shazeer and Niki Parmar and Jakob Uszkoreit and Llion Jones and Aidan N. Gomez and Lukasz Kaiser and Illia Polosukhin},
  booktitle={Neural Information Processing Systems},
  year={2017},
  url={https://api.semanticscholar.org/CorpusID:13756489}
}

@inproceedings{Loshchilov2017DecoupledWD,
  title={Decoupled Weight Decay Regularization},
  author={Ilya Loshchilov and Frank Hutter},
  booktitle={International Conference on Learning Representations},
  year={2017},
  url={https://api.semanticscholar.org/CorpusID:53592270}
}

@article{Hinton2015DistillingTK,
  title={Distilling the Knowledge in a Neural Network},
  author={Geoffrey E. Hinton and Oriol Vinyals and Jeffrey Dean},
  journal={ArXiv},
  year={2015},
  volume={abs/1503.02531},
  url={https://api.semanticscholar.org/CorpusID:7200347}
}

@inproceedings{reimers-gurevych-2019-sentence,
    title = "Sentence-{BERT}: Sentence Embeddings using {S}iamese {BERT}-Networks",
    author = "Reimers, Nils  and
      Gurevych, Iryna",
    editor = "Inui, Kentaro  and
      Jiang, Jing  and
      Ng, Vincent  and
      Wan, Xiaojun",
    booktitle = "Proceedings of the 2019 Conference on Empirical Methods in Natural Language Processing and the 9th International Joint Conference on Natural Language Processing (EMNLP-IJCNLP)",
    month = nov,
    year = "2019",
    address = "Hong Kong, China",
    publisher = "Association for Computational Linguistics",
    url = "https://aclanthology.org/D19-1410/",
    doi = "10.18653/v1/D19-1410",
    pages = "3982--3992",
}

@inproceedings{Radford2018ImprovingLU,
  title={Improving Language Understanding by Generative Pre-Training},
  author={Alec Radford and Karthik Narasimhan},
  year={2018},
  url={https://api.semanticscholar.org/CorpusID:49313245}
}

@article{Hendrycks2016GaussianEL,
  title={Gaussian Error Linear Units (GELUs)},
  author={Dan Hendrycks and Kevin Gimpel},
  journal={arXiv: Learning},
  year={2016},
  url={https://api.semanticscholar.org/CorpusID:125617073}
}

@misc{jones2019kernelbasedtranslationsconvolutionalnetworks,
      title={Kernel-based Translations of Convolutional Networks}, 
      author={Corinne Jones and Vincent Roulet and Zaid Harchaoui},
      year={2019},
      eprint={1903.08131},
      archivePrefix={arXiv},
      primaryClass={stat.ML},
      url={https://arxiv.org/abs/1903.08131}, 
}

@inproceedings{Mahowald2026EfficientAO,
  title={Efficient Analysis of the Distilled Neural Tangent Kernel},
  author={Jamie Mahowald and Brian Bell and Alex Ho and Michael Geyer},
  year={2026},
  url={https://api.semanticscholar.org/CorpusID:285541139}
}

@book{rkhs,
author = {Berlinet, Alain and Thomas-Agnan, Christine},
year = {2004},
month = {01},
pages = {},
title = {Reproducing Kernel Hilbert Space in Probability and Statistics},
isbn = {978-1-4613-4792-7},
doi = {10.1007/978-1-4419-9096-9}
}

\appendix

\newpage

\appendix
\section{Kernel Implementation and STL Semantics}\label{apd:first}

\subsection{Kernel Implementation}

The semantic inner product defined in Section \ref{subsec:kernel_def} involves an integral over the signal space $\mathcal{T}$ with respect to the measure $\mu_0$. In practice, this distance in the semantic space is approximated via Monte Carlo sampling over a set of $n$ stochastic signals $\{ \varepsilon_1, ..., \varepsilon_n\} \sim \mu_0$:
\[
\langle \rho(\varphi_i, \cdot), \, \rho(\varphi_j, \cdot) \rangle_{\mu_0} \approx \frac{1}{n} \sum_{t=1}^n \rho(\varphi_i, \varepsilon_t) \cdot \rho(\varphi_j, \varepsilon_t)
\]

To ensure scale invariance, the original formulation considers the cosine similarity of the robustness features in place of their plain dot product. Let $||\cdot||$ denote the $L_2$-norm in the Hilbert space induced by $\mu_0$; the normalized similarity $k'(\cdot, \cdot)$ is computed as:
\[
k'(\varphi_i, \varphi_j) = \frac{\langle \rho(\varphi_i, \cdot), \, \rho(\varphi_j, \cdot) \rangle_{\mu_0}}{|| \rho(\varphi_i, \cdot)|| \cdot || \rho(\varphi_j, \cdot) ||} \approx \frac{\sum_{t=1}^n \rho(\varphi_i, \varepsilon_t) \cdot \rho(\varphi_j, \varepsilon_t)}{\sqrt{\sum_{t=1}^n \rho(\varphi_i, \varepsilon_t)^2} \cdot \sqrt{\sum_{t=1}^n \rho(\varphi_j, \varepsilon_t)^2}}
\]

Finally, the similarity is projected into a Radial Basis Function (RBF) space through an exponential transformation. This maps the angular distance between formulae into a continuous spectrum $k(\cdot, \cdot) \in [0, 1]$:
\[
k(\varphi_i, \varphi_j) = \exp \Biggl( - \frac{2-2 \cdot k'(\varphi_i, \varphi_j)}{2 \sigma^2} \Biggr)
\]
The bandwidth parameter $\sigma^2$ (set by default to 0.2) controls the resolution of the semantic manifold. Lower values of $\sigma^2$ increase the "sharpness" of the kernel, assigning high similarity scores only to formulae that exhibit near-identical behaviour over the sampled trajectories.

\subsection{Boolean and Quantitative Semantics of STL}

Let $\xi: [0,T] \to \mathbb{R}^n$ be a trajectory and $\varphi$ an STL formula. STL admits two complementary semantics:

\paragraph{Boolean semantics} 
The formula $\varphi$ is satisfied by $\xi$ at time $t$ if $\xi \models_t \varphi$. For an atomic predicate $\mu = x_i \ge \theta$:
\[
\xi \models_t \mu \quad \Leftrightarrow \quad x_i(t) \ge \theta.
\]
Boolean connectives and temporal operators are interpreted in the standard way:
\[
\begin{aligned}
\xi \models_t \neg \varphi &\Leftrightarrow \xi \not\models_t \varphi,\\
\xi \models_t \varphi_1 \wedge \varphi_2 &\Leftrightarrow \xi \models_t \varphi_1 \text{ and } \xi \models_t \varphi_2,\\
\xi \models_t \varphi_1 \vee \varphi_2 &\Leftrightarrow \xi \models_t \varphi_1 \text{ or } \xi \models_t \varphi_2,\\
\xi \models_t \mathbf{F}_{[a,b]} \varphi &\Leftrightarrow \exists t' \in [t+a, t+b]: \xi \models_{t'} \varphi,\\
\xi \models_t \mathbf{G}_{[a,b]} \varphi &\Leftrightarrow \forall t' \in [t+a, t+b]: \xi \models_{t'} \varphi.
\end{aligned}
\]

\paragraph{Quantitative semantics (robustness)} 
The robustness $\rho(\varphi, \xi, t) \in \mathbb{R}$ measures the degree of satisfaction:
\[
\rho(\mu, \xi, t) = x_i(t) - \theta, \quad 
\rho(\neg \varphi, \xi, t) = -\rho(\varphi, \xi, t),
\]
\[
\rho(\varphi_1 \wedge \varphi_2, \xi, t) = \min(\rho(\varphi_1, \xi, t), \rho(\varphi_2, \xi, t)),
\quad
\rho(\varphi_1 \vee \varphi_2, \xi, t) = \max(\rho(\varphi_1, \xi, t), \rho(\varphi_2, \xi, t)),
\]
\[
\rho(\mathbf{F}_{[a,b]} \varphi, \xi, t) = \max_{t' \in [t+a, t+b]} \rho(\varphi, \xi, t'), \quad
\rho(\mathbf{G}_{[a,b]} \varphi, \xi, t) = \min_{t' \in [t+a, t+b]} \rho(\varphi, \xi, t').
\]

We write $\rho(\varphi, \xi) := \rho(\varphi, \xi, 0)$ when evaluating at $t=0$. Positive robustness implies satisfaction, negative robustness implies violation, and the magnitude quantifies the distance from the satisfaction boundary.

\section{Dataset Construction}\label{apd:third}

To ensure the training and testing datasets effectively stress the encoder’s recursive understanding, we employed an object-based augmentation strategy acting directly on the Signal Temporal Logic (STL) Abstract Syntax Tree (AST). 

\paragraph{Semantic-preserving recursive augmentation pipeline}
The core of the strategy is a stochastic priority cascade operating on AST nodes. For each node, a transformation is selected based on the probability distribution $P(\cdot)$ detailed in Table \ref{tab:augmentation_rules}. To ensure structural complexity, the pipeline implements a forcing loop: if a generated variant does not reach \emph{at least the target minimum depth} (set to $d \geq 5$), the augmentation is re-attempted until the complexity threshold is met.

\begin{table}[ht]
\centering
\caption{Semantic-preserving augmentation rules. Notation: $\Box$ (always), $\diamondsuit$ (eventually), $\mathcal{U}$ (until).}
\label{tab:augmentation_rules}
\renewcommand{\arraystretch}{1.3}
\begin{tabular}{l r l}
\toprule
\textbf{Rule} & \textbf{$P(\cdot)$} & \textbf{Mathematical Transformation} \\ 
\midrule
\texttt{Not}-injection & 0.1\% & $\varphi \to \neg \neg \varphi \quad (\text{if parent} \neq \neg)$ \\
De Morgan rewriting & 9.9\% & $(A \wedge B) \to \neg(\neg A \vee \neg B) \quad (\text{and dual})$ \\
Time partitioning & 35.0\% & $\text{op}_{I} \varphi \to \text{op}_{I_1} (\text{op}_{I_2} \varphi), \quad \text{op} \in \{\Box, \diamondsuit\}$ \\
\texttt{Until} nesting & 25.0\% & $\varphi \to (\varphi \, \mathcal{U} \, \psi) \quad \text{or} \quad (\psi \, \mathcal{U} \, \varphi)$ \\
Temporal identity & 5.0\% & $\varphi \to \text{op}_{[0,0]} \varphi, \quad \text{op} \in \{\Box, \diamondsuit\}$ \\
Distributivity & 15.0\% & $\Box_{I} (A \wedge B) \to \Box_{I} A \wedge \Box_{I} B \quad (\text{or dual})$ \\
Predicate inversion & 8.0\% & $x_i \leq \theta \to \neg(x_i > \theta)$ \\
\textit{No change (default)} & 2.0\% & $\varphi \to \varphi$ \\
\bottomrule
\end{tabular}
\end{table}

\paragraph{Semantic-altering perturbations}
The pipeline generates variants through a \textbf{stratified approach}. While some variants only undergo structural complication, 65\% of the dataset is subjected to semantic shifts that move the satisfaction boundaries. These shifts are applied orthogonally to the structural transformations: a formula may first be logically complicated to reach the target depth and subsequently perturbed numerically. Perturbations are equally split ($P=0.5$) between:

\begin{itemize}
    \item \textbf{Vibration:} multiplicative noise affecting thresholds ($\pm 10\%$) and time bounds ($W' \in [0.6 \cdot W, \, 1.8 \cdot W]$).
    \item \textbf{Shift:} additive offsets ($\Delta \in [-6, 6]$ for thresholds, $\Delta \in [-15, 40]$ for time bounds).
\end{itemize}

This layering ensures that semantically distinct formulae still retain the high syntactic complexity required to stress the encoder.

\paragraph{Post-serialization refinement}
Once serialized, three final operations de-correlate surface syntax from logic and ensure formula integrity:

\begin{itemize}
    \item \textbf{Duality shift (40\% probability):} Temporal operators are replaced with their negated duals via regex-based substitution: $\Box_{[I]} A \iff \neg \diamondsuit_{[I]} \neg A$.
    \item \textbf{Interval consistency:} The algorithm strictly enforces the temporal constraint $R > L$. Following any perturbation, the upper bound is reset as $R' = L' + W'$, where the new width $W'$ is sampled stochastically.
    \item \textbf{Empirical validation:} Every formula variant is evaluated against a stochastic dummy signal. This ensures the formula is compatible with the signal's temporal horizon and does not result in undefined robustness values or empty traces due to excessive nesting or parameter shift.
\end{itemize}

\section{Training metrics in detail}\label{apd:fourth}

In our setting, \textbf{kernel alignment} quantifies the global agreement between the semantic similarity induced by the STL kernel matrix $K \in \mathbb{R}^{B \times B}$ and the neural similarity matrix $S \in \mathbb{R}^{B \times B}$ obtained from the learned embeddings.

Denoting by $\text{vec}(\cdot)$ the flattening operator, we define:
$$
\text{kernel\_alignment}(K, S) =
\frac{\text{vec}(K) \cdot \text{vec}(S)}
{\|\text{vec}(K)\|_2 \cdot \|\text{vec}(S)\|_2}.
$$

This quantity corresponds to the cosine similarity between the two vectorized Gram matrices. An alignment value of $1.0$ flags that the learned embedding preserves all pairwise inner products up to a global scaling factor, meaning that the neural representation approximates the geometry of the STL semantic Hilbert space\footnote{While \cite{wang2022understandingcontrastiverepresentationlearning} defines alignment over discrete positive pairs, our formulation generalizes the concept to a continuous similarity structure, reflecting the graded notion of semantic agreement induced by STL robustness.}.

\textbf{Uniformity} measures how the representations are distributed on the unit hypersphere and serves as a safeguard against representational collapse. Even if pairwise similarities are locally preserved, the embedding may degenerate by concentrating most formulae in a restricted region of the space, reducing the expressive capacity of the representation.

Following \cite{wang2022understandingcontrastiverepresentationlearning}, uniformity is defined as the logarithm of the average Gaussian potential over all pairs in the batch:
$$
\mathcal{U}(Z) =
\log\!\left(
\mathbb{E}_{i,j\in B}
\left[\exp(-2\|z_i-z_j\|_2^2)\right]
\right).
$$

This formulation corresponds to the case $t=2$ in \cite{wang2022understandingcontrastiverepresentationlearning}. High uniformity indicates that the embedding occupies the hypersphere broadly, ensuring that semantically distinct specifications are mapped to separable regions of the latent space and preventing many-to-one semantic collapse.
Its values range in the extremes $[-4, 0]$ and can be briefly interpreted as follows: 
\begin{table}[h]
\centering
\label{tab:uniformity_bounds}
\begin{tabular}{cll}
\hline
\textbf{$\mathcal{U}(Z)$} & \textbf{geometrically} & \textbf{interpretation} \\ \hline
$0.0$ & complete collapse & zero discriminative power; all formulae are isomorphic \\
$-1.0$ & dense clustering & poor resolution; distinct logics occupy overlapping regions \\
$-2.5$ & reasonable distribution & reliable separation of standard temporal properties \\
$-3.0$ & good sparsity & maximum entropy; preserves fine-grained semantic nuances. \\ 
$-4.0$ & asymptotic limit & theoretical maximum spread for $d \to \infty$ with $t=2$ \\ \hline
\end{tabular}
\end{table}

% sembra inglese parlato, sistema
The \textbf{upper bound} means complete dimensional collapse since it only happens whether all the embeddings exhibit zero euclidean distance, i.e. 
$$
|| z_i - z_j ||_2^2 = 0, \quad \forall i, j \in B
$$
This would cause the uniformity to be null:
$$
\mathcal{U}(Z) = \log \left( \mathbb{E}_{i, j \in B} \left[ \exp(-2 \cdot 0^2) \right] \right)= \log \left( \mathbb{E}_{i, j \in B} \left[ \exp(0) \right] \right) = \log(1) = 0 
$$
thus we can only observe zero uniformity when the model is unable to discriminate between different formulae, neither in syntactical or in semantic terms.

On the opposite, the \emph{theoretical} \textbf{lower bound} corresponds to a perfectly uniform distribution of points across the hypersphere. Formally, for an infinite number of formulae distributed uniformly according to the surface area measure $\sigma_d$, the uniformity becomes:
$$
\mathcal{U}(Z) = \log \int_{\mathcal{S}^{d-1}} \int_{\mathcal{S}^{d-1}} \exp \left( -2 \cdot ||u - v ||_2^2 \right)  d \sigma_d(u) d \sigma_d (v)
$$
where $||u - v||_2^2 = ||u||_2^2 + ||v||_2^2 - 2 u^T v = 2 - 2 u^T v$ since $u, v \in \mathcal{S}^{d-1}$. Thus, the expression reduces to:
$$
\mathcal{U}(Z) = -2 \cdot 2 + \log \mathbb{E}_{u, v \sim \sigma_d} \left[ \exp \left( 2 \cdot 2 u^T v  \right) \right]
$$
As the dimension $d \rightarrow \infty$, $u^T v \rightarrow 0$, leading the (empirical) lower bound to:
$$
\mathcal{U}(Z) = -4 + \log \mathbb{E}_{u, v \sim \sigma_d} \left[ \exp(0) \right] = -4 + \log (1) = -4
$$

%\begin{figure}[ht!]
%    \centering
%    \includegraphics[width=1\linewidth]{all_metrics_final.png}
%    \caption{Evolution of training (top row) and validation (bottom row) metrics over approximately 30,000 training steps for the evaluated pooling strategies (\texttt{[CLS]}, mean, and \texttt{[BOS]}). The left and center panels illustrate the steady minimization of the geometric alignment loss and the corresponding increase in kernel alignment ($>0.9$), demonstrating that the encoder effectively internalizes the target semantic geometry. Importantly, validation metrics closely mirror training performance, confirming strong generalization to unseen formulae without syntactic overfitting. The right panels show the uniformity score stabilizing around $-2.4$ on the validation set, indicating that the representations maintain a well-distributed hyperspherical latent space, avoiding dimensional collapse. Across all configurations, \texttt{[CLS]} pooling (in blue) consistently exhibits the fastest and most stable convergence.}
%    \label{fig:all_training_metrics}
%\end{figure}

\section{Time and memory details}\label{apd:fifth}
The benchmarks reported in Tables~\ref{tab:embedding_totalmem_allB} and~\ref{tab:similarity_totalmem_allB} were collected to compare the STL kernel and the Transformer-based model. Both approaches were evaluated on the same dataset of STL formulae and synthetic signals, ensuring comparability. Memory usage accounts for both CPU RAM and GPU VRAM, capturing peak consumption during embedding and similarity computation stages.

The selection of formulae is deterministic: for each benchmark run, the first $B$ elements of the dataset are taken and used for computation. No random sampling is performed at this stage, ensuring reproducibility across runs. Each formula is then processed either through the Transformer-based embedding pipeline or via the STL kernel approach. For the \emph{Transformer} pipeline, each formula is tokenized using its tokenizer and forwarded through the trained model to extract embeddings. Formulae were tokenized in batches using a multithreaded DataLoader, and embeddings were extracted via forward passes without gradient computation. Similarity matrices were computed using normalized inner products of embeddings. Memory usage is monitored separately for RAM and VRAM, and timing is measured using GPU-synchronized timers to ensure precise measurement of both the forward pass and similarity computations. Benchmarks are repeated for two scenarios: one where the model is already loaded, and one including the full model loading. For the \emph{STL kernel} approach, synthetic signals are first sampled to produce $N$ trajectories, each with $1000$ points and $3$ variables. Formulae are then evaluated over these signals to compute their robustness vectors. The kernel matrix is finally computed as a similarity measure between the robustness vectors. Both memory (RAM + VRAM) and time are recorded, capturing peak consumption during robustness computation and kernel evaluation.

All experiments are run on the same hardware setup (an NVIDIA A100 80GB GPU), ensuring a fair comparison between the two methods. Memory usage includes both CPU RAM and GPU VRAM allocations, measured using system-level queries and PyTorch memory tracking. Timing accounts for all GPU synchronization points to avoid underestimating the actual execution time.

\begin{table}[!ht]
\centering
\caption{Embedding times (s) and total memory usage (MB) for STL kernel and Transformer. Total memory = RAM + VRAM. ``Loaded" refers to Transformer with model already initialized; ``Full pipeline" includes model loading.}
\label{tab:embedding_totalmem_allB}
\begin{tabular}{|c|c|cc|cccc|}
\hline
$B$ & $N$ & \multicolumn{2}{c|}{STL Kernel} & \multicolumn{4}{c|}{Transformer} \\
\hline
    &    & $T_\mathrm{Emb}$ & Mem & $T_\mathrm{Emb}$ (Loaded) & Mem (Loaded) & $T_\mathrm{Emb}$ (Full) & Mem (Full) \\
\hline
500   & 500   & 0.61  & 2059  & 0.62 & 2136 & 2.06 & 2914 \\
500   & 1000  & 0.83  & 3061  & 0.62 & 2128 & 1.98 & 2916 \\
500   & 2000  & 1.25  & 5051  & 0.62 & 2128 & 2.05 & 2918 \\
500   & 4000  & 2.21  & 8955  & 0.63 & 2134 & 2.02 & 2918 \\
500   & 8000  & 4.53  & 16995 & 0.62 & 2127 & 1.98 & 2918 \\
500   & 16000 & 12.44 & 32442 & 0.62 & 2126 & 1.99 & 2918 \\
\hline
1000  & 500   & 1.09  & 3027  & 1.11 & 2139 & 2.53 & 2911 \\
1000  & 1000  & 1.54  & 4960  & 1.11 & 2138 & 2.54 & 2911 \\
1000  & 2000  & 2.40  & 8829  & 1.11 & 2138 & 2.53 & 2910 \\
1000  & 4000  & 4.63  & 16505 & 1.11 & 2139 & 2.48 & 2910 \\
1000  & 8000  & 9.10  & 32008 & 1.11 & 2139 & 2.47 & 2909 \\
1000  & 16000 & 24.70 & 62338 & 1.11 & 2138 & 2.56 & 2910 \\
\hline
2000  & 500   & 2.18  & 4907  & 2.17 & 2158 & 3.59 & 2914 \\
2000  & 1000  & 3.03  & 8737  & 2.16 & 2151 & 3.56 & 2914 \\
2000  & 2000  & 4.77  & 16464 & 2.18 & 2155 & 3.59 & 2914 \\
2000  & 4000  & 8.36  & 31696 & 2.15 & 2156 & 3.52 & 2915 \\
2000  & 8000  & 17.08 & 62384 & 2.16 & 2152 & 3.55 & 2912 \\
2000  & 16000 & 48.86 & 123384& 2.17 & 2143 & 3.56 & 2913 \\
\hline
4000  & 500   & 4.33  & 8755  & 4.45 & 2163 & 5.91 & 2992 \\
4000  & 1000  & 6.09  & 16452 & 4.41 & 2162 & 5.84 & 2988 \\
4000  & 2000  & 9.47  & 31897 & 4.42 & 2163 & 5.88 & 2985 \\
4000  & 4000  & 17.06 & 62408 & 4.38 & 2162 & 5.86 & 2986 \\
4000  & 8000  & 34.30 & 123788& 4.42 & 2158 & 5.84 & 2980 \\
4000  & 16000 & 232.63& 246151& 4.46 & 2157 & 6.20 & 2980 \\
\hline
\end{tabular}
\end{table}

% Pairwise similarity benchmark

\begin{table}[!ht]
\centering
\caption{Pairwise similarity computation times (s) and total memory usage (MB) for STL kernel and Transformer. Total memory = RAM + VRAM. ``Loaded" refers to Transformer with model already initialized; ``Full pipeline" includes model loading. Kernel hits OOM for large $N$.}
\label{tab:similarity_totalmem_allB}
\begin{tabular}{|c|c|c|c|cc|cc|}
\hline
$B$ & $N$ & \multicolumn{2}{c|}{STL Kernel} & \multicolumn{2}{c|}{Transformer (Loaded)} & \multicolumn{2}{c|}{Transformer (Full)} \\
\hline
    &    & $T_\mathrm{Sim}$ & Mem & $T_\mathrm{Sim}$ & Mem & $T_\mathrm{Sim}$ & Mem \\
\hline
500   & 500   & 1.58  & 2059  & 0.73 & 2136 & 2.04 & 2914 \\
500   & 1000  & 2.50  & 3061  & 0.73 & 2128 & 1.99 & 2916 \\
500   & 2000  & 4.38  & 5051  & 0.73 & 2128 & 2.02 & 2918 \\
500   & 4000  & 8.21  & 8955  & 0.73 & 2134 & 2.02 & 2918 \\
500   & 8000  & 16.81 & 16995 & 0.73 & 2127 & 1.99 & 2918 \\
500   & 16000 & 40.24 & 32442 & 0.73 & 2126 & 1.99 & 2918 \\
\hline
1000  & 500   & 3.64  & 3027  & 1.21 & 2139 & 2.52 & 2911 \\
1000  & 1000  & 6.17  & 4960  & 1.22 & 2138 & 2.53 & 2911 \\
1000  & 2000  & 11.05 & 8829  & 1.21 & 2138 & 2.52 & 2910 \\
1000  & 4000  & 21.57 & 16505 & 1.22 & 2139 & 2.47 & 2910 \\
1000  & 8000  & 42.86 & 32008 & 1.22 & 2139 & 2.45 & 2909 \\
1000  & 16000 & 98.51 & 62338 & 1.21 & 2138 & 2.56 & 2910 \\
\hline
2000  & 500   & 10.54  & 4907  & 2.29 & 2158 & 3.53 & 2914 \\
2000  & 1000  & 18.33  & 8737  & 2.28 & 2151 & 3.59 & 2914 \\
2000  & 2000  & 33.87  & 16464 & 2.29 & 2155 & 3.62 & 2914 \\
2000  & 4000  & 66.37  & 31696 & 2.27 & 2156 & 3.53 & 2915 \\
2000  & 8000  & 132.29 & 62384 & 2.28 & 2152 & 3.54 & 2912 \\
2000  & 16000 & 434.57 & 123384& 2.29 & 2143 & 3.55 & 2913 \\
\hline
4000  & 500   & 32.65  & 16776  & 4.59 & 2163 & 5.85 & 2992 \\
4000  & 1000  & 60.11  & 32112  & 4.55 & 2162 & 5.93 & 2988 \\
4000  & 2000  & 114.98 & 62653  & 4.53 & 2155 & 6.05 & 2985 \\
4000  & 4000  & 227.25 & 123651 & 4.51 & 2156 & 6.47 & 2986 \\
4000  & 8000  & 552.04 & 245724 & 4.58 & 2152 & 5.83  & 2980 \\
4000  & 16000 & 1089.28$^*$& 489898$^*$ & 4.65 & 2152 & 6.01  & 2980 \\
\hline
\end{tabular}
\end{table}

\newpage

\begin{figure}[!h]
    \centering
    \includegraphics[width=0.7\linewidth]{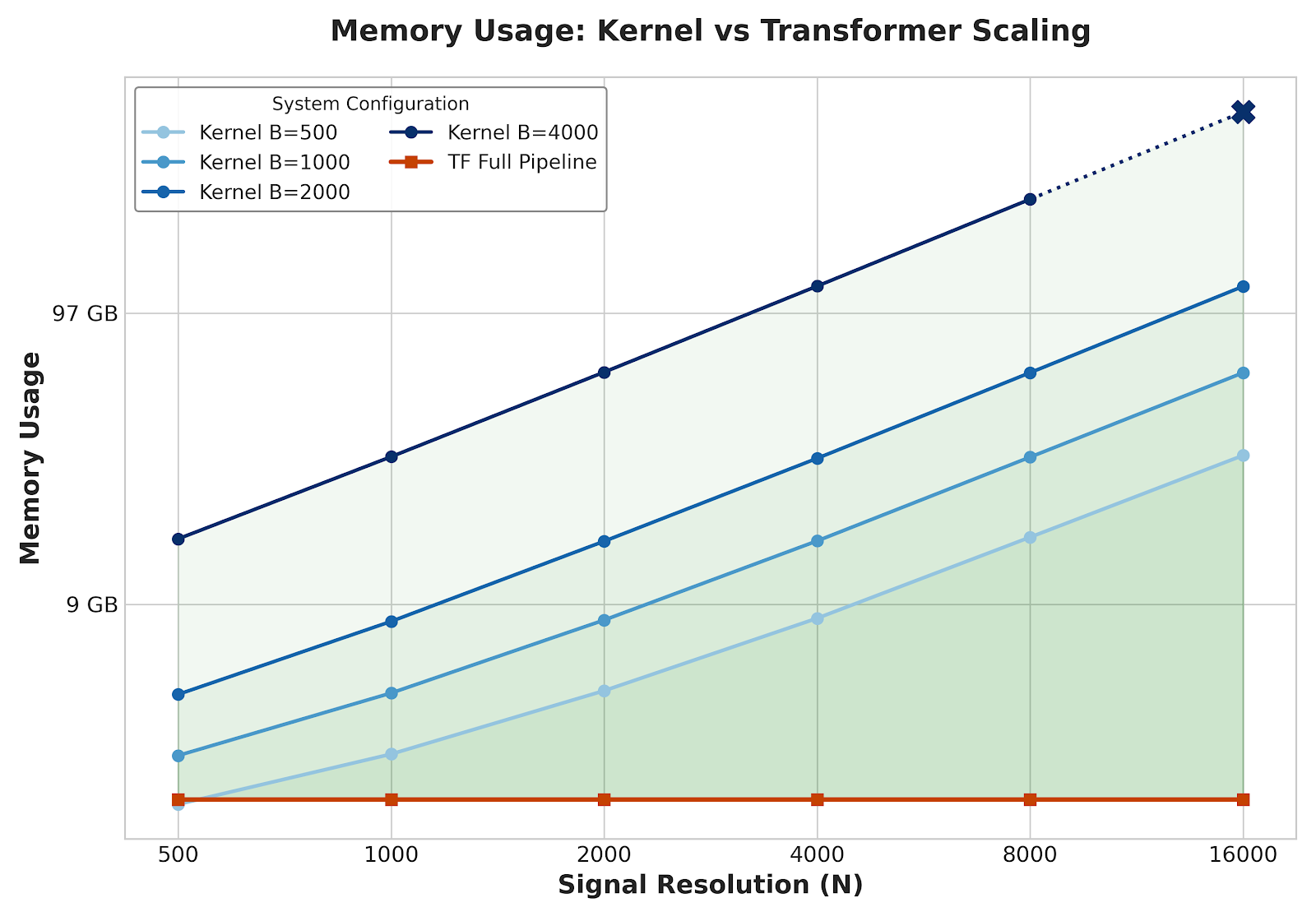}   \label{fig:pairsim-time}
\end{figure}

\begin{figure}[!h]
    \centering
    \includegraphics[width=0.7\linewidth]{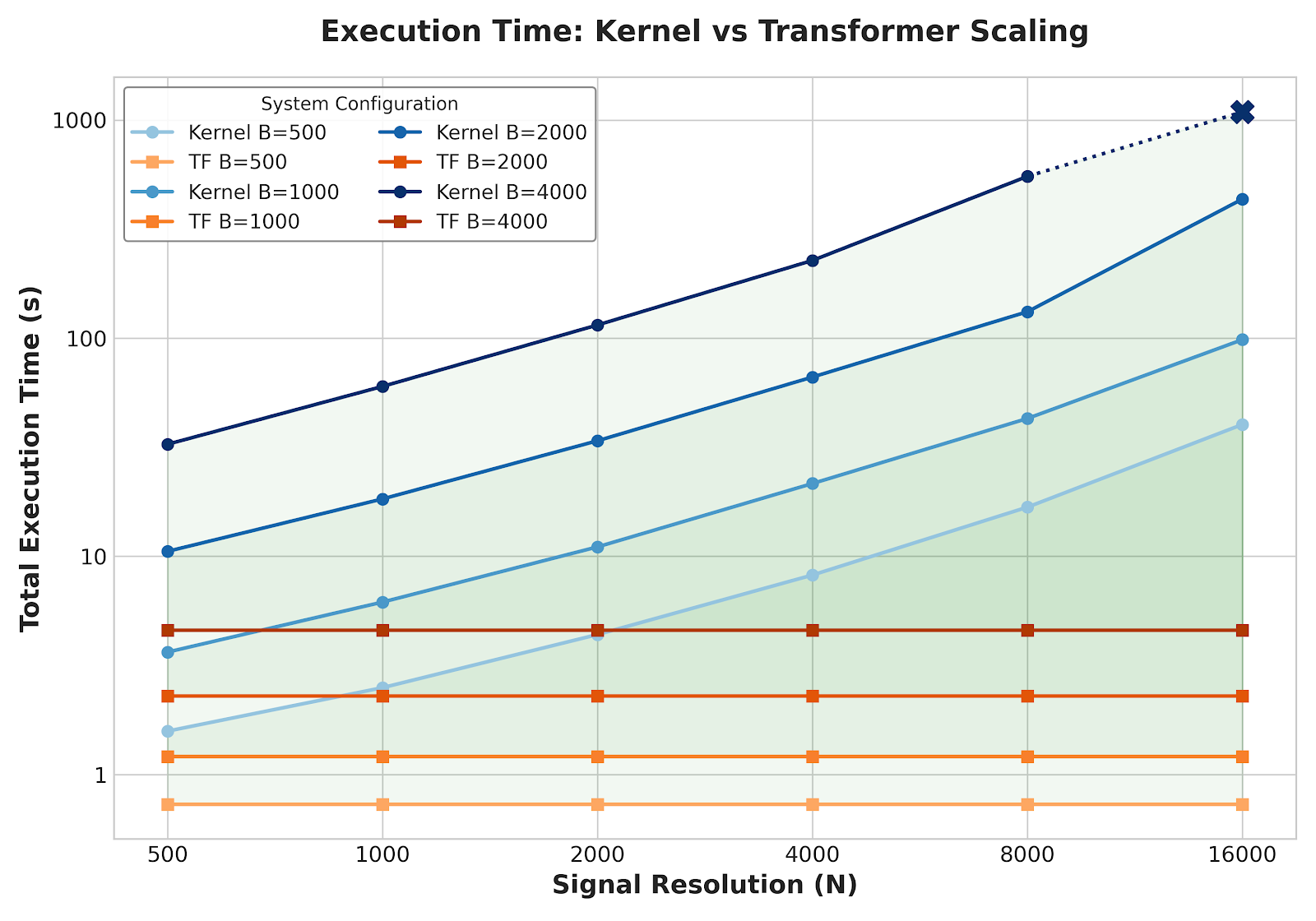}
    \label{fig:pairsim-mem}
\end{figure}

Computational scaling of the kernel method and the Transformer-based embedding pipeline. Runtime (Figure \ref{fig:pairsim-time}) and memory usage (Figure \ref{fig:pairsim-mem}) are reported as functions of signal resolution $N$ and batch size $B$. Projected values (dashed) indicate configurations exceeding memory limits (OOM).

\newpage

\section{\texttt{[CLS]}, \texttt{[BOS]} and mean pooling comparison}\label{apd:gradient_flow}

In Section \ref{sec:methodology}, we introduced three distinct pooling strategies to aggregate the token-level representations produced by the Transformer encoder into a single fixed-dimensional embedding. To evaluate the sensitivity of our distillation framework to the chosen aggregation method, we present a detailed semantic understanding evaluation for each strategy. 

Table \ref{tab:semantic_results_cls} (reproduced here from the main text for ease of comparison), Table \ref{tab:semantic_results_mean}, and Table \ref{tab:semantic_results_arch_v1} report the performance of \texttt{[CLS]}, mean, and \texttt{[BOS]} pooling, respectively. Across all metrics, the results demonstrate that the proposed kernel-alignment objective is highly robust to the choice of pooling operation. 

While \texttt{[CLS]} pooling exhibits a marginally tighter alignment with the target kernel similarities (yielding the lowest Mean Absolute Error of 0.034 on equivalent pairs), both mean and \texttt{[BOS]} pooling successfully internalize the continuous semantic geometry. All three strategies consistently and effectively separate semantically equivalent formulae from non-equivalent and syntactically similar ones. This confirms that the architectural bottleneck effectively compresses logical meaning and filters out syntactic noise, regardless of the specific token aggregation heuristic employed.

\begin{table}[h!]
\centering
\caption{Semantic understanding evaluation for \texttt{[CLS]} pooling (reference).}
\begin{tabular}{lccc}
\hline
\textbf{Metric} & \textbf{Equivalent} & \textbf{Non-equivalent} & \textbf{Syntactically similar} \\
\hline
Neural similarity & 0.966 & 0.182 & 0.308 \\
Kernel similarity & 0.997 & 0.170 & 0.225 \\
MAE (semantic gap) & 0.034 & 0.072 & 0.112 \\
\hline
Relative neural distance & 0.137 & 0.833 & 0.770 \\
Relative kernel distance & 0.105 & 0.830 & 0.775 \\
\hline
\end{tabular}
\label{tab:semantic_results_cls_app}
\end{table}

\begin{table}[h!]
\centering
\caption{Semantic understanding evaluation for mean pooling.}
\begin{tabular}{lccc}
\hline
\textbf{Metric} & \textbf{Equivalent} & \textbf{Non-equivalent} & \textbf{Syntactically similar} \\
\hline
Neural similarity & 0.954 & 0.201 & 0.332 \\
Kernel similarity & 0.997 & 0.163 & 0.221 \\
MAE (semantic gap) & 0.044 & 0.066 & 0.143 \\
\hline
Relative neural distance & 0.169 & 0.833 & 0.749 \\
Relative kernel distance & 0.100 & 0.837 & 0.779 \\
\hline
\end{tabular}
\label{tab:semantic_results_mean}
\end{table}

\begin{table}[h!]
\centering
\caption{Semantic understanding evaluation for \texttt{[BOS]} pooling.}
\begin{tabular}{lccc}
\hline
\textbf{Metric} & \textbf{Equivalent} & \textbf{Non-equivalent} & \textbf{Syntactically similar} \\
\hline
Neural similarity        & 0.946 & 0.171 & 0.280 \\
Kernel similarity        & 0.997 & 0.152 & 0.182 \\
MAE (semantic gap)       & 0.052 & 0.070 & 0.133 \\
\hline
Relative neural distance & 0.186 & 0.834 & 0.794 \\
Relative kernel distance & 0.111 & 0.848 & 0.818 \\
\hline
\end{tabular}
\label{tab:semantic_results_arch_v1}
\end{table}

\end{document}